\RequirePackage{fix-cm}
\documentclass[twocolumn]{svjour3}          % twocolumn
\smartqed  % flush right qed marks, e.g. at end of proof
\usepackage{graphicx}
\usepackage{color, soul}
\usepackage{transparent}
\usepackage{import}
\usepackage{rotating}
\usepackage{booktabs}
\usepackage[table]{xcolor}
\usepackage{tabularx}
\usepackage{tabulary}
\usepackage[english, status=draft]{fixme}
\usepackage[linesnumbered, ruled]{algorithm2e}
\fxusetheme{color}
\usepackage{amsmath}
\usepackage{amssymb}
\usepackage{subfigure}
\usepackage{pifont}
\usepackage{multirow}
\usepackage{lscape}
\usepackage{scrextend}
\usepackage{hyperref}
\usepackage{mathrsfs}
\usepackage{enumitem}
\newlength{\colwidth}
\newcommand{\nrsfm}[0] {{\sc nrs{\rm\em f}\sc m}}

\definecolor{darkolivegreen}{rgb}{0, 0.5, 0}
\newcommand{\rev}[1]{{\color{darkolivegreen} {#1}}}

\begin{document}

\title{A Benchmark and Evaluation of \\ Non-Rigid Structure from Motion%\thanks{Grants or other notes
}
\subtitle{}


\author{Sebastian Hoppe Nesgaard Jensen, \\
Mads Emil Brix Doest, \\
Henrik Aan\ae s, \\
Alessio Del Bue
}


\institute{
        Sebastian Hoppe Nesgaard Jensen \at 
        DTU Compute, Denmark\\
        \email{snje@dtu.dk} 
    \and
        Mads Emil Brix Doest \at 
        DTU Compute, Denmark\\
        \email{mebd@dtu.dk} 
    \and
        Henrik Aan\ae s \at 
        DTU Compute, Denmark\\
        \email{aanes@dtu.dk} \\
        \emph{Last author with equal contribution.}
    \and
        Alessio Del Bue \at 
       Pattern Analysis and Computer Vision (PAVIS) \\
       Visual Geometry and Modelling (VGM) Lab\\
       Istituto Italiano di Tecnologia (IIT), Genova, 08028, Italy.  \\
       \email{alessio.delbue@iit.it}\\
       \emph{Last author with equal contribution.}
}

\date{Received: date / Accepted: date}

\maketitle

\begin{abstract}
Non-Rigid structure from motion (\nrsfm{}), is a long standing and central problem in computer vision and its solution is necessary for obtaining 3D information from multiple images when the scene is dynamic. A main issue regarding the further development of this important computer vision topic, is the lack of high quality data sets. We here address this issue by presenting a data set created for this purpose, which is made publicly available, and considerably larger than the previous state of the art. To validate the applicability of this data set, and provide an investigation into the state of the art of \nrsfm{}, including potential directions forward, we here present a benchmark and a scrupulous evaluation using this data set. This benchmark evaluates 18 different methods with available code that reasonably spans the state of the art in sparse \nrsfm{}. This new public data set and evaluation protocol will provide benchmark tools for further development in this challenging field.
\end{abstract}

\section{Introduction}

The estimation of structure from motion (SfM) using a monocular image sequence is one of the central problems in computer vision. This problem has received a lot of attention, and truly impressive advances have been made over the last ten to twenty years \cite{Hartley:Zisserman:book2000,szeliski2010computer,ozyecsil2017survey}. It plays a central role in robot navigation, self-driving cars, and 3D reconstruction of the environment, to mention a few. A central part of maturing regular SfM is the availability of sizeable data sets with rigorous evaluations, e.g. \cite{Menze2015CVPR}\cite{aanaesinteresting}.
 
The regular SfM problem, however, primarily deals with rigid objects, which is somewhat at odds with the world we see around us. That is, trees sway, faces express themselves in various expressions, and organic objects are generally non-rigid. The issue of making this obvious and necessary extension of the SfM problem is referred to as the non-rigid structure from motion problem (\nrsfm{}). A problem that also has a central place in computer vision. The solution to this problem is, however, not as mature as the regular SfM problem. A reason for this is certainly the intrinsic difficulty of the problem and the scarcity of high quality data sets and accompanying evaluations. Such data and evaluations allow us to better understand the problem domain and better determine what works best and why.
 
To address this issue, we here introduce a high quality data set, with accompanying ground truth (or reference data to be more precise) aimed at evaluating non-rigid structure from motion. To the best of our knowledge, this data set is significantly larger and more diverse than what has previously been available -- c.f.\ Section~\ref{sec:dataset} for a comparison to previous evaluations of \nrsfm{}. The presented data set better capture the variability of the problem and gives higher statistical strength of the conclusions reached via it. Accompanying this data set, we have conducted an evaluation of 18 state of the art methods, hereby validating the suitability of our data set, and providing insight into the state of the art within \nrsfm{}. This evaluation was part of the competition we held at a CVPR 2017 workshop, and still ongoing. It is our hope and belief that this data set and evaluation will help in furthering the state of the art in \nrsfm{} research, by providing insight and a benchmark. The data set is publicly available at {\url{http://nrsfm2017.compute.dtu.dk/dataset}} together with the description of the evaluation protocol.

This paper is structured by first giving an overview of the \nrsfm{} problem, followed by a general description of related work, wrt.\ other data sets. This section is then followed by a presentation of our data set, including an overview of the design considerations, c.f.\ Section~\ref{sec:dataset}, which is followed by a presentation of our proposed protocol for evaluation, c.f.\ Section~\ref{sec:evaluation_matrix}. This leads to the result of our benchmark evaluation in Sections~\ref{sec:evaluation}. The paper is rounded off by a discussion and conclusions in Section~\ref{sec:discussion_and_conclusion}.

%-----------------------------------------------------------------------------------
\section{The \nrsfm{} Problem} \label{sec:NRSFMProblem}

In this section, we will provide a brief introduction of the \nrsfm{} problem, followed by a more detailed overview of the ways this problem has been addressed. The intention is to establish a taxonomy to base our experimental design and evaluation upon. In particular, we review  sparse NRSfM methods as these approaches are the one evaluated in our benchmark. 

The standard/rigid SfM problem, c.f.\ e.g.\ \cite{Hartley:Zisserman:book2000}, is an inverse problem aimed at finding the camera positions (and possibly internal parameters) as well as 3D structure -- typically represented as a static 3D point set, $Q$ -- from a sequence of 2D images of a rigid body. The 2D images are typically reduced to a sparse set of tracked 2D point features, corresponding to the 3D point set, $Q$. The most often employed observation model, linking 2D image points to 3D points and camera motion is either the \emph{perspective camera model}, or the \emph{weak perspective} approximation hereof. The weak perspective camera model is derived from the full perspective model, by simplifying the projective effect of 3D point depth, i.e. the distance between the camera and 3D point.

The extension from rigid structure from motion to the non-rigid case is by allowing the 3D structure, here points $\mathbf{Q}_f$, to vary from frame to frame, i.e.
\begin{align}
\mathbf{Q}_f    &=    \begin{bmatrix}
                \mathbf{Q}_{f, 1} & \mathbf{Q}_{f, 2} & \cdots & \mathbf{Q}_{f, P}
            \end{bmatrix}
            \enspace ,
\end{align}
Where $\mathbf{Q}_{f, p}$ is the 3D position of point $p$ at frame $f$. To make this \nrsfm{} problem  well-defined, a prior or regularization is often employed. Here most of the cases target the spatial and temporal variations of $\mathbf{Q}_f$. The fitness of the prior to deformation in question is a crucial element in successfully solving the \nrsfm{} problem, and a main difference among \nrsfm{} methods is this prior.

In this study, we denote \nrsfm{} methods according to a three category taxonomy, i.e.\ the \textbf{deformable model} used (statistical or physical), the \textbf{camera model} (affine, weak or full perspective) and the ability to deal with \textbf{missing data}. The remainder of this section will elaborate this taxonomy by relating it with the current literature, leading up to a discussion of how the \nrsfm{} methods we evaluate, c.f.\ Table~{\ref{table:methods}}, span the state of the art.

\subsection{Deformable Models}
The description of our taxonomy will start with the underlying structure deformation model category, divided into statistical and physical based models.

\subsubsection{Statistical}

This set of algorithms apply a statistical deformation model with no direct connection to the physical process of structure deformations. They are in general heuristically defined a priori to enforce constraints that can reduce the ill-posedness of the \nrsfm{} problem. The most used low-rank model in the \nrsfm{} literature falls into this category, utilizing the assumption that 3D deformations are well described by linear subspaces (also called basis shapes). The low-rank model was first introduced almost 20 years ago by Bregler et al. \cite{Bregler:etal:CVPR2000} solving \nrsfm{} through the formalisation of a factorization problem, as analogously proposed by Tomasi and Kanade for the rigid case \cite{Tomasi:Kanade:IJCV92}. However, strong nonlinear deformations, such as the one appearing in articulated shapes, may drastically reduce the effectiveness of such models. Moreover, the first low-rank model presented in \cite{Bregler:etal:CVPR2000} acted mainly as a constraint over the spatial distribution of the deforming point cloud and it did not restrict the temporal variations of the deforming object.   

Differently, Gotardo and Martinez. \cite{GotardoPAMI2011} had the intuition to use the very same DCT bases to model camera and deformation motion instead, assuming those factors are smooth in a video sequence. This approach was later expanded on by explicitly modeling a set of complementary rank-3 spaces, and to constrain the magnitude of deformations in the basis shapes~{\cite{GotardoCVPR2011}}. An extension of this framework, increased the generalization of the model to non-linear deformations, with a kernel transformation on the 3D shape space using radial basis functions~{\cite{GotardoICCV2011}}. This switch of perspective addressed the main issue of increasing the number of available DCT bases, allowing more diverse motions, while not restricting the complexity of deformations. Later, further extension and optimization have been made to low-rank and DCT based approaches.
Valmadre and Lucey \cite{ValmadreCVPR2012} noticed that the trajectory should be a low-frequency signal, thus laying the ground for an automatic selection of DCT basis rank via penalizing the trajectory's response to one or more high-pass filters. Moreover, spatio-temporal constraints have been imposed both for temporal and spatial deformations \cite{akhter2012bilinear}. 

A related idea proposed by Li et al. \cite{li2018structure} attempts at grouping recurrent deformations in order to better describe deformations. At its core, the method has an additional clustering step that links together similar deformations.  
Recently a new prior model, related to the Kronecker-Markov structure of the covariance of time-varying 3D point, very well generalizes several priors introduced previously \cite{simon2017kronecker}. Another recent improvement is given by Ansari et al.'s usage of DCT basis in conjunction with singular value thresholding for camera pose estimation~{\cite{scalesurface2017}}.
  
Similar spatial and temporal priors have been introduced as regularization terms while optimizing a cost function solving for the \nrsfm{} problem, mainly using a low-rank model only. Torresani et al.\ \cite{TorresaniPAMI2008} proposed a probabilistic PCA model for modelling deformations by marginalizing some of the variables, assuming Gaussian distributions for both noise and deformations. Moreover, in the same framework, a linear dynamical model was used to represent the deformation at the current frame as a linear function of the previous. Brand \cite{Brand:Bhotika:CVPR2001} penalizes deformations over the mean shape of the object by introducing  sensible parameters over the degree of flexibility of the shape. Del Bue et al. \cite{DelBue:etal:AMFG2005} instead compute a more robust non-rigid factorization, using a 3D mean shape as a prior for \nrsfm{} \cite{DelBue:IJCV2013}. In a non-linear optimization framework, Olsen et al. \cite{olsen2008implicit} include  $l_2$ penalties both on the frame-by-frame deformations and on the closeness of the reconstructed points in 3D given their 2D projections. Of course, penalty costs introduce a new set of hyper-parameters that weights the terms, implying the need for  further tuning, that can be impracticable when cross-validation is not an option. Regularization has also been introduced in formulations of Bundle Adjustment for \nrsfm{} \cite{aanaes2002estimation} by including smoothness deformations via $l_2$ penalties mainly \cite{DelBue:etal:IVC2007} or constraints over the rigidity of pre-segmented points in the measurement \cite{DelBueCVPR2006}.

Another important statistical principal is enforcing that low-rank bases are independent. In the coarse to fine approach of Bartoli et al. \cite{BartoliCVPR2008}, base shapes are computed sequentially by adding the basis, which explains most of the variance in respect to the previous ones. They also impose a stopping criteria, thus, achieving the automatic computation of the overall number of bases. The concept of basis independence clearly calls for a statistical model close to Independent Component Analysis (ICA). To this end, Brandt et al. \cite{Brandt:2009} proposed a prior term to minimize the mutual information of each basis in the \nrsfm{} model. 
Low-rank models are indeed compact but limited in the expressiveness of complex deformations, as noted in \cite{zhu2014complex}. To solve this problem, Zhu et al. \cite{zhu2014complex} use  a temporal union of subspace that associate at each cluster of frames in time a specific subspace. Such association is solved by adopting a cost function promoting self-expressiveness \cite{elhamifar2013sparse}. Similarly, both spatial and temporal union of subspaces was used also to account for independently deforming multiple shapes \cite{agudo2017dust,MultiBody2017}. Interestingly, such union of subspaces strategy was previously adopted to solve for the multi-body 3D reconstruction of independently moving objects \cite{zappella2013joint}. Another option is  to use an over-complete representation of subspaces that can still be used by imposing sparsity over the selected bases \cite{compressible2016}.  In this way, 3D shapes in time can have a compact representation, and they can be theoretically characterized as a block sparse dictionary learning problem. In a similar spirit, Hamsici et al.\ propose to use the input data for learning spatially smooth shape weights using rotation invariant kernels~{\cite{gotardo:ECCV2012}}.  

All these approaches for addressing \nrsfm{} with a low-rank model have provided several non-linear optimization procedures, mainly using Alternating Least Squares (ALS), Lagrange Multipliers and alternating direction method of multipliers (ADMM). Torresani et al. first proposed to alternate between the solution of camera matrices, deformation parameters and basis shapes. This first initial solution was then extended by Wang et al. \cite{Wang:2008} by constraining the camera matrices to be orthonormal at each iteration, while Paladini et al. \cite{DelBue:Agapito:IJCV2011} strictly enforced the matrix manifold of the camera matrices to increase the chances to converge to the global optimum of the cost function. All these methods were not designed to be strictly convergent, for this reason, a Bilinear Augmented Multiplier Method (BALM) \cite{DelBue:etal:PAMI2012} was introduced to be convergent while implying all the problems constraints being satisfied. Furthermore, robustness in terms of outlying data was then included to improve results in a proximal method with theoretical guarantees of convergence to a stationary point \cite{wang2015practical}. 

Despite the non-linearity of the problem, it is possible to relax the rank constraint with the trace norm and solve the problem with convex programming. Following this strategy, Dai et al. provided  one of the first effective closed form solutions to the low-rank problem \cite{DaiIJCV2014}. Although their convex solution, resulting from relaxation, did not provide the best performance, a following iterative optimization scheme gave improved results. In this respect, Kumar et al. proposed a further improvement on their previous approach, where deformations are represented as a spatio-temporal union of subspaces rather than a single subspace~{\cite{MultiBody2017}}. Thus complex deformation can be represented as the union of several simple ones as already described in the previous paragraphs. To notice that evaluation is performed with synthetic generated data only. 

Later Kumar \cite{kumar2020non} proposed a set of improvements over Dai et al. approach \cite{DaiIJCV2014}. Namely, metric rectification was performed using incomplete information by choosing arbitrarily a triplet of solutions among the one available. The solution in \cite{kumar2020non} proposes a method to select the best among the available triplets using a rotation smoothness heuristic as a decision criteria. Then, a further improvement is algorithmic. Instead of using Dai et al. strategy with a matrix shrinkage operator that equally penalizes all the singular values, the method in \cite{kumar2020non} introduces a weighted nuclear norm function during optimisation. More recently Ornhag et al. \cite{ornhag2020unified} proposed a unified optimization framework for low-rank inducing penalties that can be readily applied to solve  for \nrsfm{}. The main advantage of the approach is the ability to combining bias reduction in the estimation and nonconvex low-rank inducing objectives in the form of a weighted nuclear norm. 

On the one hand, the Procrustean Normal Distribution (PND) model was proposed as an effective way to implicitly separate rigid and non-rigid deformations \cite{Lee:PAMI:2017,8052164}. This separation provides a relevant regularization, since rigid motion can be used to obtain a more robust camera estimation, while deformations are still sampled as a normal distribution as done similarly previously \cite{TorresaniPAMI2008}. Such a separation is obtained by enforcing an alignment between the reconstructed 3D shapes at every frame. This should in practice factor out the rigid transformations from the statistical distribution of deformations. The PND model has been then extended to deal with more complex deformations and longer sequences \cite{cho2016complex}.

\subsubsection{Physical}

Physical models represent a less studied class wrt. NRSfM, which should ideally be the most accurate for modelling \nrsfm{}. Of course, applying the right physical model requires a knowledge of the deformation type and object material, which is
information not readily available a priori.

A first class of physical models assume that the non-rigid object is a piecewise partition into parts, i.e.\ a collection of pre-defined or estimated patches that are mostly rigid or slightly deformable. This observation is certainly true for objects with articulated deformations, as it naturally models natural and mechanical shapes connected into parts. One of the first approaches to use this strategy is given by Varol et al. \cite{Varol09a}. By preselecting a set of overlapping patches from the 2D image points, and assuming each patch is rigid, homography constraints can be imposed at each patch, followed by global 3D consistency being enforced  using the overlapping points. However, the rigidity of a  patch, even if small, is a very hard constraint to impose and it does not generalise well for every non-rigid shape. Moreover, dense point-matches over the image sequence are required to ensure a set of overlapping points among all the patches. A relaxation to the piece-wise rigid constraint was given by Fayad et al. \cite{Fayad:etal:2010}, assuming each patch deforming with a quadratic physical model, thus, accounting for linear and bending deformations. These methods all require an initial patch segmentation and the number of overlapping points, to this end, Russel et al.\ \cite{RussellCVPR2011} optimize the number of patches and overlap by defining an energy based  cost function. This approach was further extended and generalised to deal with general videos \cite{russell2014video} and energy functional that includes temporal smoothing  \cite{Golyanik_2019}. The method of Lee et al. \cite{concensus2016} instead use 3D reconstructions of multiple combinations of patches and define a 3D consensus  between a set of patches. This approach provides a fast way to bypass the segmentation problem and robust mechanism to prune out wrong local 3D reconstructions. The method was further improved to account for higher degrees of missing data in the chosen patches so to generalise better the capabilities of the approach in challenging \nrsfm{} sequences \cite{8778692}.

Differently from these approaches, Taylor et al. \cite{TaylorCVPR2010} constructs a triangular mesh, connecting all the points, and considering each triangle as being locally rigid. Global consistency is here imposed to ensure that the vertexes of each triangle coincide in 3D. Again, this approach is to a certain extent similar to \cite{Varol09a}, which requires a dense set of points in order to comply with the local rigidity constraint.

A strong prior, which helps dramatically to mitigate the ill-posedness of the problem, is obtained by considering the deformation isometric, i.e.\ the metric length of curves does not change when the shape is subject to deformations (e.g. paper and metallic materials to some extent). A first solution considering a regularly sampled surface mesh model was presented in \cite{salzmann2007surface}. Using an assumption that a surface can be approximated as infinitesimally planar, Chhatkuli et al.\ \cite{chhatkuli2014non}  proposed a local  method that frame \nrsfm{} as the solution of Partial Differential Equations (PDE) being able to deal with missing data as well. As a further update \cite{parashar2017isometric} formalizes the framework in the context of Riemannian geometry, which led to a practical method for solving the problem in linear time and scaling for a relevant number of views and points. Furthermore, a convex formulation for \nrsfm{} with inextensible deformation constraints was implemented using Second-Order Cone Programming (SOCP), leading to a closed form solution to the problem~\cite{chhatkuli2017inextensible}. Vincente and Agapito implemented soft inextensibility constraints \cite{Vicente:etal:2012} in an energy minimization framework, e.g.\ using recently introduced techniques for discrete optimization. 

Another set of approaches try to directly estimate the deformation function using high order models. {Del Bue} and Bartoli \cite{DelBueICCV2011} extended and applied 3D warps such as the thin plate spline, to the \nrsfm{} problem. Starting from an approximate mean 3D reconstruction, the warping function can be constructed and the deformation at each frame can be solved by iterating between camera and 3D warp field estimation. Finally, Agudo et al. introduced the use of Finite Elements Models (FEM) in \nrsfm{} \cite{agudo2016sequential}. As these models are highly parametrized, requiring the knowledge of the material properties of the object (e.g. the Young modulus), FEM needs to be approximated in order to be efficiently estimated, however, in ideal conditions it might achieve remarkable results, since FEM is a consolidated technique for modelling structural deformations. Lately, Agudo and Moreno-Nouger presented a duality between standard statistical rank-constrained model and a new proposed force model inspired from the Hooke's law \cite{agudo2017force}. However, in principle, their physical model can account for a wider range of deformations than rank-based statistical approaches.

\subsection{Missing Data}

The initial methods for \nrsfm{} assumed complete 2D point matches among views when observing a deformable object. However, given self and standard occlusions, this is rarely the case.  Most approaches for dealing with such missing data in \nrsfm{} were framed as a matrix completion problem, i.e.\ estimate the missing entries of the matrix storing the 2D coordinates obtained by projecting each deforming 3D point. 

Torresani et al.\ \cite{Torresani:etal:CVPR2001} first proposed removing rows and lines of the matrix corresponding to missing entries in order to solve the \nrsfm{} problem. However, this strategy suffers greatly from even small percentages of missing data, since the subset of completely known entries can be very small. 
Most of the iterative approaches indeed include an update step of the missing entries \cite{DelBue:Agapito:IJCV2011,DelBue:etal:PAMI2012} where the missing entries become an explicit unknown to estimate. Gotardo et al.\ \cite{GotardoPAMI2011} instead strongly reduce the number of parameters by estimating only the camera matrix explicitly under severe missing data. This variable reduction is known as VARPRO in the optimization literature. It has been recently revisited in relation to several structure from motion problems \cite{Hong_2017_CVPR}. 

\subsection{Camera Model}

Most \nrsfm{} methods in the literature  
assume a weak perspective camera model. However, in cases where the object is close to the camera and undergoing strong changes in depth, time-varying perspective distortions can significantly affect the measured 2D trajectories.  

As low-rank \nrsfm{} is treated as a factorization problem, a straightforward extension is to follow best practices from rigid SfM for perspective camera.  Xiao and Kanade \cite{Xiao:Kanade:ICCV2005} have developed a two step factorization algorithm for reconstruction of 3D deformable shapes under the full perspective camera model. This is done using the assumption that a set of basis shapes are known to be independent. Vidal and Abretske \cite{vidal2006nonrigid} have also proposed an algebraic solution to the non-rigid factorization problem. Their approach is, however, limited to the case of an object being modelled with two independent basis shapes and viewed in five different images. Wang et al.\ \cite{wang2007structure} proposed a method able to deal with the perspective camera model, but under the assumption that its internal calibration is already known. They update the solutions from a weak perspective to a full perspective projection by refining the projective depths recursively, and then refine all the parameters in a final optimization stage. Finally, Hartley and Vidal \cite{Hartley:Vidal:2008} have proposed a new closed form linear solution for the perspective camera case.
This algorithm requires the initial estimation of a multifocal tensor, which the authors report is very sensitive to noise. Llado et al.\ \cite{Llado:etal:ICPR2006,llado2010non} proposed a non-linear optimization procedure. It is based on the fact that it is possible to detect nearly rigid points in the deforming shape, which can provide the basis for a robust camera calibration.

\begin{table*}[t]
  \centering
  \caption{Methods included in our \nrsfm{} evaluation with annotations of how they fit into our taxonomy.}
  \label{table:methods}
  \rowcolors{2}{gray!20}{white}
  \begin{tabular}{@{} *6r @{}}
    \toprule
    \textbf{{Method}} & \textbf{{Citation}} & \textbf{{Deformable Model}} & \textbf{{Camera Model}} &  \textbf{{Missing Data}}\\
    \midrule
    BALM    & \cite{DelBue:etal:PAMI2012}                       &  Statistical  & Orthographic  & Yes  \\
    Bundle  & \cite{DelBue:etal:IVC2007}            &  Statistical  & Weak Perspective & Yes\\
    Compressible    & \cite{compressible2016}       & Statistical   & Weak Perspective  &  - \\
    Consensus   & \cite{concensus2016}              & Physical  &  Orthographic  &  - \\
    CSF     & \cite{GotardoPAMI2011}                & Statistical     &  Weak Perspective &  Yes \\
    CSF2    & \cite{GotardoCVPR2011}                &  Statistical    &  Orthographic  &  Yes \\
    EM PPCA & \cite{TorresaniPAMI2008}          & Statistical    & Weak Perspective  &  Yes \\
    KSTA    & \cite{GotardoICCV2011}                & Statistical   & Orthographic  &  Yes \\
    MDH     & \cite{chhatkuli2017inextensible}                        & Physical &  Perspective &  Yes \\
    MetricProj   & \cite{DelBue:Agapito:IJCV2011}          & Statistical  & Orthographic  &  Yes \\
    MultiBody    & \cite{MultiBody2017}             & Statistical   & Orthographic  &  - \\
    PTA     & \cite{Akhter:2011}                    &  Statistical  &  Orthographic &  - \\
    RIKS    & \cite{gotardo:ECCV2012}               &  Statistical   & Orthographic  &  - \\
    ScalableSurface & \cite{scalesurface2017}       & Statistical    &  Orthographic &  Yes \\ 
    SoftInext   & \cite{Vicente:etal:2012}          &  Physical  &  Perspective &  Yes \\ 
    SPFM    & \cite{DaiIJCV2014}                    & Statistical    & Orthographic  &  - \\
    CMDR & \cite{Golyanik_2019}           & Physical   & Orthographic & - \\
    F-consensus & \cite{8778692}           & Physical   & Orthographic & yes \\

    \bottomrule                 
  \end{tabular}
\end{table*}

\subsection{Evaluated Methods}
We have chosen a representative subset of the aforementioned methods, which are summarized according to our taxonomy in Table~\ref{table:methods}. This gives us a good representation of recent works, distributed according to our taxonomy with a decent span of deformation models (statistical/physical) and camera models (orthographic, weak perspective or perspective).
This also takes into account in-group variations such as DCT basis for statistical deformation and isometry for physical deformation. Even lesser used priors, such as compressibility, are represented.
While this is not a full factorial study, we think this reasonably spans the recent state of the art of \nrsfm{}. Our choice has, of course, also been influence by method availability, as we want to test the author's original implementation, to avoid our own implementation bias/errors. All in all, we have included 18 methods in our evaluation.

Note that we have chosen not to include the method of Taylor et al.~\cite{TaylorCVPR2010}, even if code is available, the approach failed approximately two thirds of the time when tested on our data set.

\section{Dataset} \label{sec:dataset}
As stated, in order to compare state of the art methods for \nrsfm{},  we have compiled a larger data set for this purpose. Even though there is a lack of empirical evidence w.r.t.\ \nrsfm{}, it does not imply, that no data sets for \nrsfm{} exist. 

As an example in ~\cite{concensus2016},~\cite{GotardoPAMI2011},~\cite{GotardoCVPR2011},~\cite{GotardoICCV2011},~\cite{MultiBody2017},~\cite{Akhter:2011},~\cite{gotardo:ECCV2012} and~\cite{DaiIJCV2014}, a combination of two data sets are used. Namely seven sequences of a human body from the CMU motion capture database~\cite{cmumocap}, two MoCap sequences of a deforming face~\cite{Torresani:etal:NIPS03,del2005non}, a computer animated shark~\cite{Torresani:etal:NIPS03} and a challenging flag sequence \cite{Fayad:etal:2010}. To the best of our knowledge, this list in Table \ref{table:dataset} represents the most used evaluation data sets for \nrsfm{} with available ground truth.

The CMU data set~\cite{cmumocap} captures the motion of humans. Since the other frequently used data sets are also related to animated faces~\cite{Torresani:etal:NIPS03,del2005non}, this implies that there is a high over representation of humans in this state of the art and that a higher variability in the deformed scenes viewed is deemed beneficial.  In addition, the shark sequence \cite{Torresani:etal:NIPS03} is not based on real images and objects but on computer graphics and pure simulation. As such, there is a need for new data sets, with reliable ground truth or reference data,\footnote{With real measurements like ours the 'ground truth' data also include noise, why 'reference data' is a more correct term.} and a higher variability in the objects and deformations used.

\begin{table*}[t]
  \centering
  \caption{A description of the previous data set sequences with available ground truth. The table shows the number of frames and points, the way to generate the sequence (mainly with motion capture data) and the type of shape used.}
  \label{table:dataset}
  \rowcolors{2}{gray!20}{white}
  \begin{tabular}{@{} *5r @{}}
    \toprule
    %\hline
    \textbf{{Name}} & \textbf{{Citation}} & \textbf{{Frames$\times$Points}} & \textbf{{Type}} & \textbf{{Shape}} \\
    \midrule
    shark   & \cite{TorresaniPAMI2008}     &  240 $\times$ 91  & Synthetic & Animal motion\\
    face1   & \cite{TorresaniPAMI2008}     &  74 $\times$ 37  & Mocap & Face motion\\
    face2   & \cite{TorresaniPAMI2008}     &  316 $\times$ 40  & Mocap & Face motion\\
    cubes   & \cite{XiaoIJCV2006}     &  200 $\times$ 14  & Synthetic & ToyProblem\\
    face\_occ   & \cite{DelBue:Agapito:IJCV2011}     &  70 $\times$ 37  & Mocap & Face motion\\
    flag        & \cite{Fayad:etal:2010} &  540 $\times$ 50  & Mocap & Cloth deformation\\
    yoga    & \cite{Akhter:2011}     &  307 $\times$ 41  & Mocap & Human motion \\
    drink   & \cite{Akhter:2011}     &  1102 $\times$ 41  & Mocap & Human motion\\
    stretch   & \cite{Akhter:2011}     &  307 $\times$ 41  & Mocap & Human motion\\
    dance   & \cite{Akhter:2011}     &  264 $\times$ 41  & Mocap & Human motion\\
    pickup   & \cite{Akhter:2011}     &  357 $\times$ 41  & Mocap & Human motion\\
    walking   & \cite{Akhter:2011}     &  260 $\times$ 41  & Mocap & Human motion\\
    capoeira   & \cite{GotardoPAMI2011}     &  250 $\times$ 41  & Mocap & Human motion\\
    jaws   & \cite{GotardoPAMI2011}     &  321 $\times$ 49  & Synthetic & Animal motion\\
    \bottomrule                 
  \end{tabular}
\end{table*}

As such, we here present a data set consisting of five widely different objects/scenes and deformations. The physical object motions are generated  mechanically using animatronics, therefore assuring experimental repeatability. Furthermore, we have defined six different camera motions using orthographic and full perspective camera models. This setup, all in all, gives $60$ different sequences organized in a factorial experimental design, thus, enabling a more stringent statistical analysis. In addition to this, since we have tight 3D surface models of our objects or scenes, we are able to determine occlusions of all 2D feature points. This in turn gives a realistic handling of missing data, which is often due to object self occlusion. Given this procedure of generating occlusions, missing data  always follow a more realistic structured pattern in contrast with the most common, and unrealistic, random process of removing 2D measurement entries used in previous evaluation dataset.

As indicated, these data sets are achieved by stop-motion using mechanical animatronics. These are recorded in our robotic setup previously used for generating high quality data sets c.f.\  e.g.\ \cite{aanaes2016large}. We will here present details of our data capture pipeline, followed by a brief outline and discussion of design considerations.

The goal of the data capturing is to produce 3 types of related data:
\begin{description}[style=multiline,leftmargin=2.75cm]
    \item[\textbf{Ground Truth:}] A series of 3D points that change over time.
    \item[\textbf{Input Tracks: }] 2D tracks used as input. 
    \item[\textbf{Missing Data:}] Binary data indicating the tracks that are occluded at specific     image frames.
\end{description}
We record the step-wise deformation of our animatronics from $K$ static views, obtaining both image data and dense 3D surface geometry. We obtain 2D point features by applying standard optical flow tracking \cite{bouguet2001pyramidal} to the image sequence obtained from each of the $K$ views, which is then reprojected onto the recorded surface geometry. The ground truth is then the union of these 3D tracks. By using optical flow for tracking instead of MoCap markers, we obtain a more realistic set of ground truth points. We create input 2D points by projecting the recorded ground truth using a virtual camera in a fully factorial design of camera paths and camera models. 

In the following, we will detail some of the central parts of the above procedure.

\subsection{Animatronics \& Recording Setup}
\label{sec:animatronics}

\begin{figure}[t]
\centering
\includegraphics[width=0.235\textwidth]{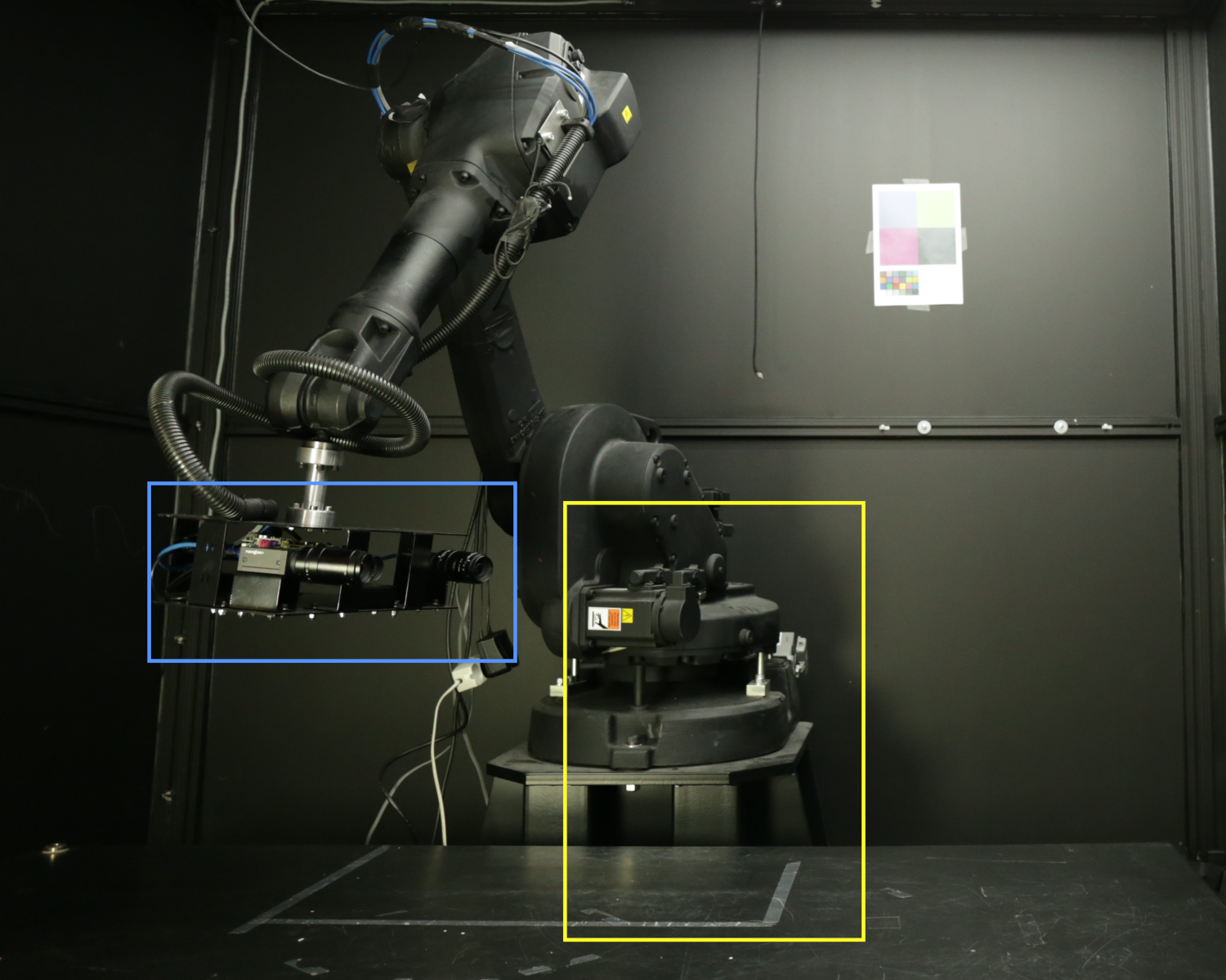}
\includegraphics[width=0.235\textwidth]{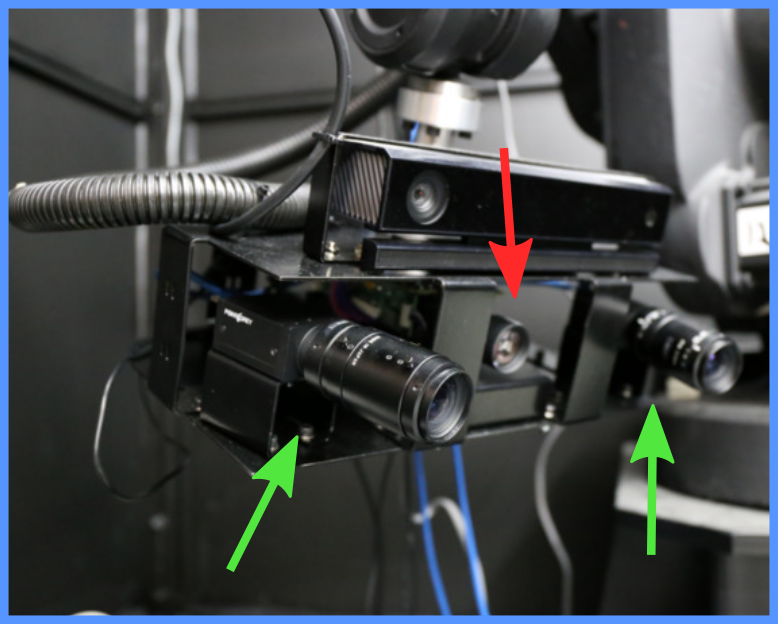}
\centering
\caption{Images of the robot cell for dataset acquisition. Left image shows the robot with the structured light scanner (blue box) and the area where the animatronic systems are positioned (yellow box). Right image shows the structured light scanner up close, green arrows show the position of the PointGrey Grasshopper3 cameras, and the red arrow marks the Lightcrafter 4500 projector.}
\label{fig:recording_setup}
\end{figure}

\begin{figure}[t]
\centering
    \subfigure[Articulated]{\includegraphics[width=0.165\textwidth]{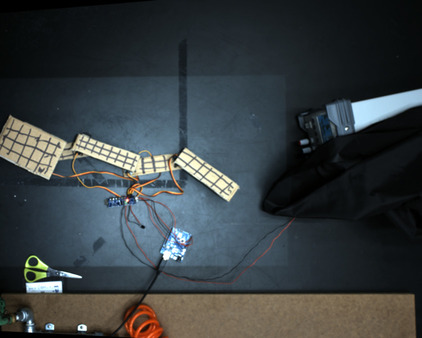}\label{fig:articulated}}
    \subfigure[Bending]{\includegraphics[width=0.165\textwidth]{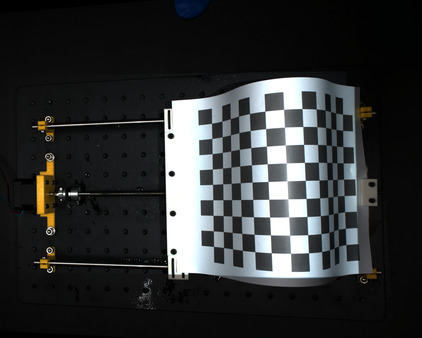}\label{fig:paperbender}}
    \subfigure[Deflation]{\includegraphics[width=0.165\textwidth]{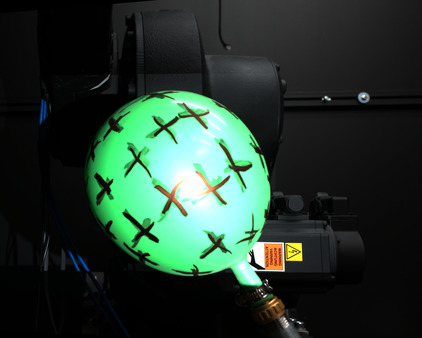}\label{fig:deflation}}
    \subfigure[Stetching]{\includegraphics[width=0.165\textwidth]{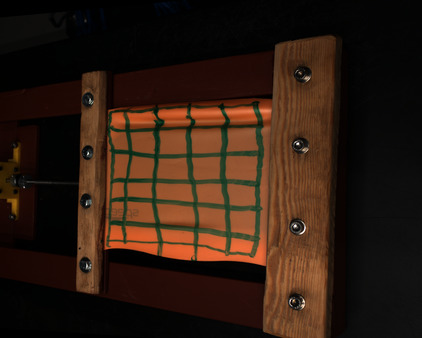}\label{fig:stretch}}
    \subfigure[Tearing]{\includegraphics[width=0.165\textwidth]{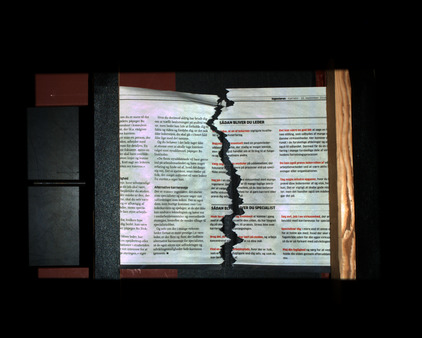}\label{fig:ripper}}
    \caption{Animatronic systems used for generating specific types of non-rigid motion.}
    \label{fig:animatronics}
\end{figure}

Our stop-motion animatronics are five mechatronic devices capable of computer controlled gradual deformation. They are shown in Fig.~\ref{fig:animatronics}, and they cover five types of deformations: Articulated Motion, Bending, Deflation, Stretching, and Tearing. We believe this covers a good range of interesting and archetypal deformations. It is noted, that \nrsfm{} has previously been tested on bending and tearing~\cite{TaylorCVPR2010,Vicente:etal:2012,chhatkuli2017inextensible,concensus2016}, but without ground truth for quantitative comparison. Additionally, elastic deformations, like deflation and stretching, are quite commonplace but did not appear in any previous data sets, to the best of our knowledge.

The animatronics can hold a given deformation or pose for a large extent of time, thus, allowing us to record accurately the object's geometry. We, therefore, do not need a real-time 3D scanner or elaborate multi-scanner setup. Instead, our recording setup consists of an in-house built structured light scanner mounted on an industrial robot as shown in Fig.~\ref{fig:recording_setup}.  This does not only provide us with accurate 3D scan data, but the robot's mobility also enables a full scan of the object at each deformation step.

The structured light scanner utilizes two PointGrey Grasshopper3 9.1MP CCD cameras and a projector WinTech Lightcrafter 4500 Pro projecting patterns onto the scene and acquiring images. Then, we use the Heterodyne Phase Shifting method \cite{PhaseShifting} to compute the point clouds using 16 periods across the image and 9 shifts. We verified precision according to standard VDI 2634-2~\cite{vdi2634}, and found that the scanner has a form error of [0.01mm, 0.32mm], a sphere distance error of [-0.33mm 0.50mm] and a flatness error of [0.29mm, 0.56mm]. This is approximately 2 orders of magnitude better than the results we see in our evaluation of the \nrsfm{} methods.

\subsection{Recording Procedure}
\label{sec:recording}

\begin{figure}[t]
\centering
\includegraphics[width=0.4\textwidth]{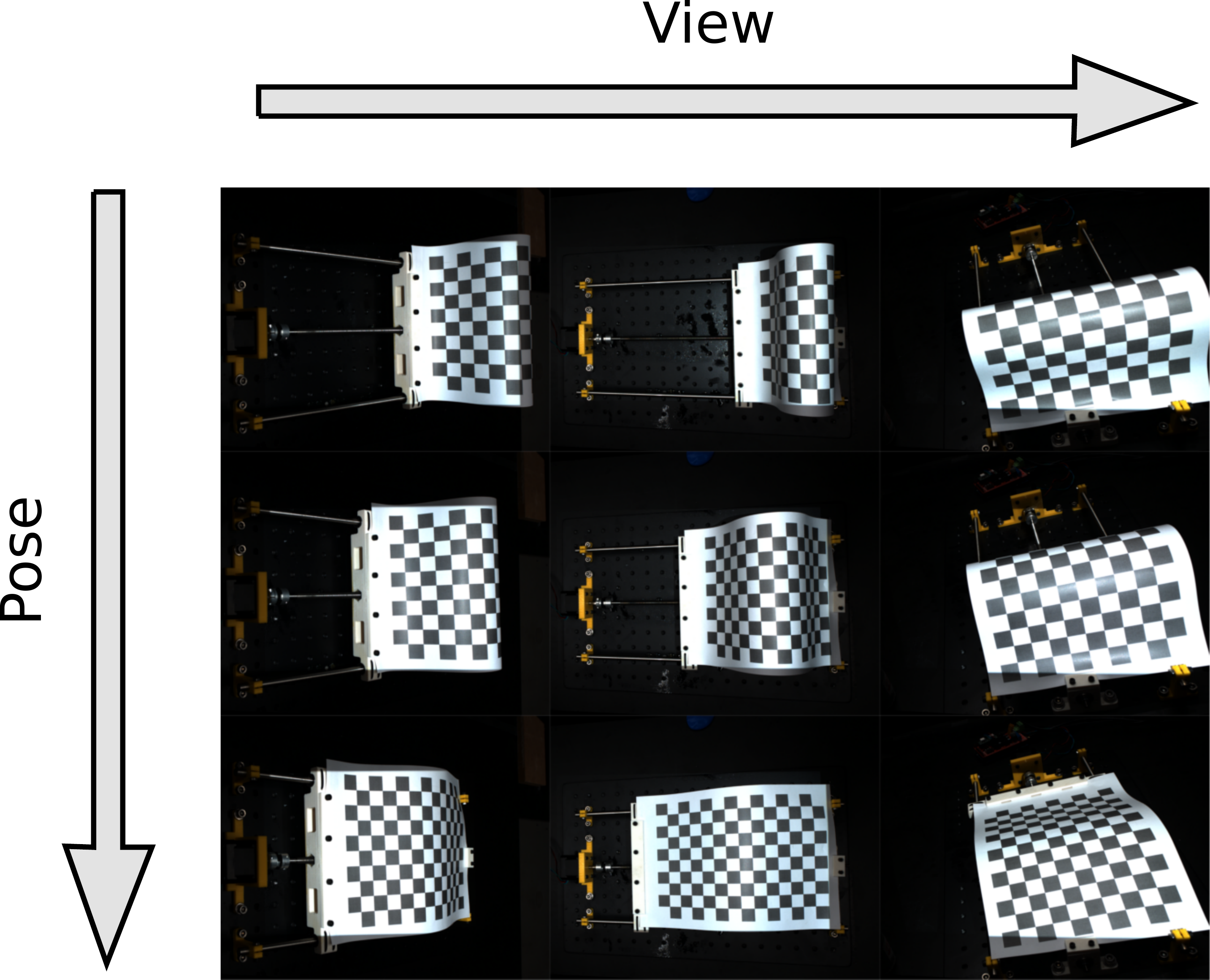}
\caption{Illustrative sample of our multi-view, stop-motion recording procedure. Animatronic pose evolves vertically and scanner view change horizontally.}
\label{fig:seq_example}
\end{figure}

The recording procedure acquires for each shape a series of image sequences and surface geometries of its deformation over $F$ frames. We record each frame from $K$ static views with our aforementioned structured light scanner. As such we obtain $K$ image sequences with $F$ images in each. We also obtain $F$ dense surface reconstructions, one for each frame in the deformation. The procedure is summarized in pseudo code in Algorithm~\ref{algo:record}. Fig.~\ref{fig:seq_example} illustrates sample images of three views obtained using the above process.

\begin{algorithm}
Let $F$ be the number of frames\\
Let $k$ be the number of static scan views $K$\\
\For{$f \in F$} {
    Deform animatronic to pose $f$\\
    \For{$k \in K$} {
        Move scanner to view $k$\\
        Acquire image $I_{f, k}$\\
        Acquire structured light scan $S_{f, k}$\\
    }
    Combine scans $S_{f, k}$ for full, dense surface $S_f$
}
\caption{Process for recording image data for tracking and dense surface geometry for an animatronic.}
\label{algo:record}
\end{algorithm}

\subsection{3D Ground Truth Data}
\label{sec:3d_reference}

The next step is to take acquired images $I_{f, k}$ and surfaces $S_f$, and extract the ground truth points.
We do this by applying optical flow  tracking~\cite{bouguet2001pyramidal} as implemented in OpenCV 2.4 to obtain 2D tracks, which are then reprojected onto $S_f$. The union of these reprojected tracks gives us the ground truth, $Q$. This process is summarized in pseudo code in Algorithm~\ref{algo:flow}.

\begin{algorithm}
Let $F$ be the number of frames\\
Let $k$ be the number of static scan views $K$\\
Let $S_f$ be the surface at frame $f$\\
Let $I_{f, k}$ be the image from view $k$, frame $f$\\
$S = \{S_1 \dotsc S_F$\}\\
\For{$k \in K$} {
    $I_k = \{I_{1, k} \dotsc I_{F, k}\}$\\
    Apply optical flow~\cite{bouguet2001pyramidal} to $I_k$ to get 2D tracks $T_k$\\
    Reproject $T_k$ onto $S$ to get 3D tracks $Q_k$
}
$Q = \{Q_1 \dotsc Q_K\}$
\caption{Process for extracting the ground truth $Q$ from recorded images and surface scans.}
\label{algo:flow}
\end{algorithm}

\begin{figure}[b]
    \centering
    \subfigure[Circle]{\includegraphics[width=0.125\textwidth]{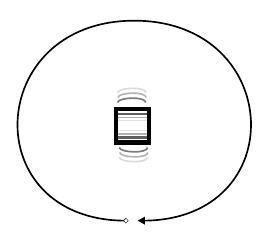}}
    \subfigure[Flyby]{\includegraphics[width=0.125\textwidth]{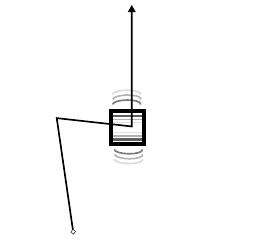}}
    \subfigure[Half Circle]{\includegraphics[width=0.125\textwidth]{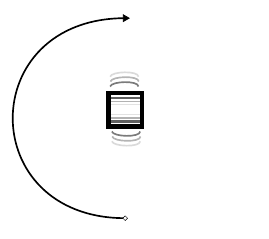}}
    \subfigure[Line]{\includegraphics[width=0.125\textwidth]{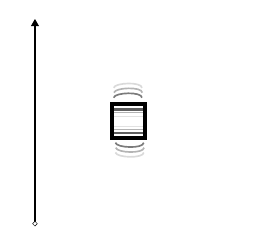}}
    \subfigure[Tricky]{\includegraphics[width=0.125\textwidth]{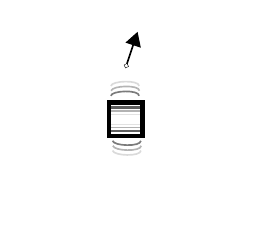}\label{fig:tricky}}
    \subfigure[Zigzag]{\includegraphics[width=0.125\textwidth]{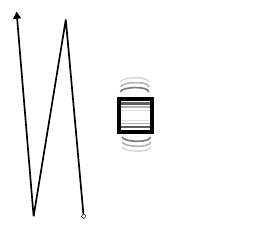}}
    \caption{Camera path taxonomy. The box represents the deforming scene and the wiggles illustrates the main direction of deformation, e.g. the direction of stretching.}
    \label{fig:camera_path}
\end{figure}

\subsection{Projection using a Virtual Camera}
\label{sec:virtual_camera}

To produce the desired input, we project the ground truth $\mathbf{Q}$ using a virtual camera, similar to what has been done in~\cite{concensus2016,GotardoPAMI2011,DaiIJCV2014,del2005non}. This step has two factors related to the camera that we wish to control for: Path and camera model. To keep our design factorial, we define six different camera paths, which will all be used to create the 2D input. They are illustrated in Fig.~\ref{fig:camera_path}. We believe these are a good representation of possible camera motion with both linear motion and panoramic panning. The Circle and Half Circle paths correspond well to the way scans are performed in SfM and structured light  methods: By moving around the target object we try to cover most of its shape. Line and Flyby are to simulate a scenario where instead the camera move linearly as in the automotive and drone-alike movements respectively. Zigzag and Tricky motions are about having depth variations in the camera movement, which is important for perspective camera, where each frame will have different projective distortions. Tricky camera path resembles more a critical motion in the direction of the optical ray of the camera as expected, for instance, in medical imaging. To conclude, as mentioned earlier, the camera model can be either orthographic or perspective.

The factorial combination of these elements yields to 12 input sequences for each ground truth. Additionally, as we have previously recorded the dense surface for each frame (see Sec.~\ref{sec:recording}), we estimate missing data via self-occlusion. Specifically, we create a triangular mesh for each $S_f$ and estimate occlusion via raycasting into the camera along the projection lines. Vertices whose ray intersects a triangle on the way to the camera are removed, from the input for the given frame, as those vertices would naturally be occluded. In this way, we ensure as realistic as possible structured missing data by modelling self-occlusion given the different camera paths. This process is summarized in pseudo code in Algorithm~\ref{algo:proj}.
\begin{algorithm}
Let $F$ be the number of frames\\
Let $P$ be the set of camera paths shown in Fig.~\ref{fig:camera_path}\\
Let $C$ be either perspective or orthographic\\
Let $Q_f$ be the ground truth at frame $f$\\
Let $S_f$ be the surface at frame $f$\\
\For{$S_f \in \{S_1 \dotsc S_F\}$}{
    Estimate mesh $M_f$ from $S_f$
}
\For{$c \in C$}{
    \For{$p \in P$}{
        \For{$f \in F$}{
            Set camera pose to $p_f$\\
            Project $Q_f$ using model $c$ to get points $w_f$\\
            Do occlusion test $q_f$ against $M_f$ to get missing data $d_f$
        }
        $W_{c, p} = \{w_1 \dotsc w_F\}$\\
        $D_{c, p} = \{d_1 \dotsc d_F\}$
    }
}
\caption{Creation of input tracks $W_{c, p}$ and missing data $D_{c, p}$ from ground truth $Q$ for each combination of camera path $p$ and model $c$.}
\label{algo:proj}
\end{algorithm}

\subsection{Discussion}
While stop-motion does allow for diverse data creation, it is not without drawbacks. Natural acceleration is easily lost when objects deform in a step-wise manner and recordings are unnaturally free of noise like motion blur. However, without this technique, it would have been prohibitive to create data with the desired diversity and accurate 3D ground truth.

The same criticism could be levied against the use of a virtual camera, it lacks the shakiness and acceleration of a real world camera. On the other hand, it allows us to precisely vary both the camera path and camera model. This enables us to perform a factorial analysis, in which we can study the effects of different configurations on \nrsfm{}. As we show in Sec.~\ref{sec:evaluation} some interesting conclusions are drawn from this analysis. Most \nrsfm{} methods are designed with an orthographic camera in mind. As such investigating the difference between data under orthographic and perspective projection is of interest. Such an investigation is only practically possible using a virtual camera.

\section{Evaluation Metric} \label{sec:evaluation_matrix}

In order to compare the  methods of Table~\ref{table:methods} w.r.t.\ our data set, a metric is needed. The purpose is to project the high dimensional 3D reconstruction error into (ideally) a one dimensional measure. 

Several different metrics have been proposed for \nrsfm{} evaluation in the past literature, e.g.\ the Frobenius norm~\cite{Paladini:etal:2009}, mean~\cite{gotardo:ECCV2012}, variance normalized mean~\cite{GotardoCVPR2011} and RMSE~\cite{TaylorCVPR2010}.

All of the above mentioned evaluation metrics are based on the $L2$-norm in one form or another. A drawback of the $L2$-norm is its sensitive to large errors, often letting a few outliers dominating the evaluation. To address this, we incorporate robustness into our metric, by introducing truncation of the individual 3D point reconstruction errors. In particular, our metric is based on a RMSE measure similar used in Taylor et al.~\cite{TaylorCVPR2010}.

Given the visualisation effectiveness and general adoption  of box plots \cite{velleman1981applications}, we propose to use their whisker function to identify and to model outliers in the error distribution. Such a strategy will  enable the inclusion of outliers in the metric with the additional benefit of reducing their influence in the RMSE.
Consider $E$ being the set of point-wise errors ($||\mathbf{X}_{f,p} - \mathbf{Q}_{f, p}||$) and $E_1$, $E_3$ as the first and third quartile of that set. As described in \cite{williamson1989box}, we define the whisker as $w=\frac{3}{2} (E_3 - E_1)$, then any point that is more than a whisker outside of the interquantile range ($IQR=E_3$ - $E_1$) is considered as an outlier. Those outliers are then truncated at $E_3 + w$ allowing them to be included in a RMSE without dominating the result. This strategy works well for approximately normally distributed data. With this in mind, our truncation function is defined as follows,
\begin{align}
  t\left(\mathbf{x}, \mathbf{q}\right) = \begin{cases}
    ||\mathbf{x} - \mathbf{q}||, & ||\mathbf{x} - \mathbf{q}|| < E_3 + w\\
    E_3 + w, & \text{otherwise}\\
  \end{cases}\label{eq:truncation}
\end{align}
Thus the robust RMSE is defined as,
\begin{align}
m\left(\mathbf{Q}, \mathbf{X}\right) &= \sqrt{\frac{1}{FP}\sum_{f,p}^{F,P}{t\left(\mathbf{X}_{f,p}, \mathbf{Q}_{f, p}\right)}}\label{eq:metric}.
\end{align}
A \nrsfm{} reconstruction is given in an arbitrary coordinate system, thus we must align the reference and reconstruction before computing the error metric. This is typically done via Procrustes Analysis~\cite{gower1975generalized}, but as it minimizes the distance between two shapes in a $L2$-norm sense it is also sensitive to outliers. Therefore, we formulate our alignment process as an optimization problem based on the robust metric of Eq.~\ref{eq:metric}. Thus the combined metric and alignment is given by,

\begin{align}
  m(\mathbf{X}, \mathbf{Q}) &=
  \min_{s, \mathbf{R}, \mathbf{t}} \sqrt{ \frac{1}{FP} \sum_{f, p} t\left(s\left[\mathbf{R}\mathbf{X}_{fp} + \mathbf{t} \right], \mathbf{Q}_{fp} \right)},\label{eq:align}\\
  \text{where}~s&=\text{scale,}\nonumber\\
  \mathbf{R}&=\text{rotation and reflection,}\nonumber\\
  \mathbf{t}&=\text{translation.}\nonumber
\end{align}

An implication of using a robust, as opposed to a $L2$-norm, is that the minimization problem of \eqref{eq:align} cannot be achieved by a standard Procrustes alignment, as done in \cite{TaylorCVPR2010}. As such, we optimize \eqref{eq:align} using the Levenberg-Marquardt method, where $s$, $\mathbf{R}$ and $\mathbf{t}$ have been initialized via Procrustes alignment~\cite{gower2004procrustes}. 
In summary, \eqref{eq:align} defines the alignment and metric that has been used for the evaluation presented in Sec.~\ref{sec:evaluation}. 

Notice also that this registration procedure estimates a single rotation and translation for the entire sequence. In this way, we avoid the practise of registering the GT 3D shape at every frame of the reconstructed 3D sequence. Such frame-by-frame procedure does not account for the global temporal consistency of the reconstructed 3D sequence and in particular regarding possible sign flips of the 3D shape, scale variations, or reflections that might happen abruptly from one frame to the other during reconstruction. Registering the 3D ground truth frame-by-frame is also unrealistic, because in general, it is not feasible to do in a real operative reconstruction scenario where 3D GT is not available.

To conclude, the choice of an evaluation metric always has a streak of subjectivity and for this reason, we investigated the sensitivity of choosing a particular one. We did this by repeating our evaluation with another robust metric, where the minimum track-wise distance between the ground truth and reconstruction was used. By just using the n-th percentile, instead of our truncation, the magnitude of the RMSE significantly decreases, but the major findings and conclusions, as presented in Sec.~\ref{sec:evaluation}, were the same. As such we conclude that our conclusions are not overly sensitive to the choice of metric. 

\section{Evaluation}
\label{sec:evaluation}
With our data set and robust error metric, we have performed a thorough evaluation and analysis of the state-of-the-art in \nrsfm{}, which is presented in the following. This is done in part as an explorative analysis and in part to answer some of what we see as most pressing, open questions in \nrsfm{}. Specifically:

\begin{itemize}
  \item Which algorithms perform the best?
  \item Which deformable models have the best performance or generalization?
  \item How well can the state-of-the-art handle data from a perspective camera?
  \item How well can the state-of-the-art handle occlusion-based missing data?
\end{itemize}

To answer these questions, we perform our analysis in a factorial manner, aligned with the factorial design of our data set. To do this, we view a \nrsfm{} reconstruction as a function of the following factors:

\begin{description}[style=multiline,leftmargin=3.4cm]
    \item[\textbf{Algorithm} $a_i$:] Which algorithm was used.
    \item[\textbf{Camera Model} $m_j$:] Which camera model was used (perspective or orthographic).
    \item[\textbf{Animatronics} $s_k$:]  Which animatronics sequence was reconstructed.
    \item[\textbf{Camera Path} $p_l$:]  How the camera moved.
    \item[\textbf{Missing Data} $d_n$:] Whether occlusion based missing data was used.
\end{description}
We design our evaluation to be almost fully crossed, meaning we obtain a reconstruction for every combination of the above factors. 

The only missing part is that the authors of MultiBody~\cite{MultiBody2017} only submitted reconstructions for orthographic camera model.

Our factorial experimental design allows us to employ a classic statistical method known as ANalysis Of VAriance (ANOVA)~\cite{seber2012linear}. The ANOVA not only allow us to deduce the precise influence of each factor on the reconstruction but also allows for testing their significance.
To be specific, we model the reconstruction error in terms of the following bilinear model,

\begin{align}
  y = \mu& + a_i + m_j + s_k + p_l + d_n\label{eq:anova} \\
          &+ as_{ik} + ap_{il} + ad_{in} + ms_{jk} \nonumber\\
          & + mp_{jl} + md_{jn} + sp_{kl} + sd_{kn} + pd_{ln}, \nonumber
\end{align}
where,
\begin{align*}
  y         &= \text{reconstruction error},\\
  \mu       &= \text{overall average error},\\
  xy_{i,j}  &= \text{interaction term between factor $x_i$ and $y_j$}.
\end{align*}

This model, Eq. \eqref{eq:anova}, contains both linear and interaction terms, meaning the model reflects both factor influence as independent and as cross effects, e.g. $as_{ik}$ is the interaction term for 'algorithm' and 'animatronics'. For each term, we test for significance by choosing between two hypotheses:
\begin{align}
\mathcal{H}_0&: c_0 = c_1 = \dotsc = c_N\\
\mathcal{H}_1&: c_0 \neq c_1 \neq \dotsc \neq c_N\nonumber
\end{align}
with $c_n$ being a term from \eqref{eq:anova} e.g. $a_i$ or $md_{jn}$. Typically, $\mathcal{H}_0$ is referred to as the null hypothesis, meaning the term $c_n$ has no significant effect. ANOVA allows for estimating the probability of falsely rejecting the null hypothesis for each factor. This statistic is referred to as the p-value. A term is referred to as being statistically significant if its p-value is below a certain threshold. In this paper we consider a significance threshold of 0.0005 or approximately $3.5\sigma$. As such, we clearly evaluated which factors are important for \nrsfm{} and which are not. 

Another interesting property of the ANOVA is that all coefficients in a given factor sums to zero,
\begin{align}
  \sum_{i = 0}^N c_i = 0.
\end{align}
So each factor can be seen as adjusting the predicted reconstruction error from the overall average. It should be noted that the ``algorithm''/``camera model'' interaction $am_{ij}$ has been left out of \eqref{eq:anova} due to MultiBody~\cite{MultiBody2017} only being tested with one camera model.

The error model of \eqref{eq:anova} is not directly applicable to the error of all algorithms as not all state-of-the-art methods from Table~\ref{table:methods} can deal with missing data. As such we perform the evaluation in two parts. One where we disregard missing data and include all available methods from Table~\ref{table:methods}, and one where we use the subset of methods that handles missing data and utilize the full model of \eqref{eq:anova}. The former is covered in Sec.~\ref{sec:evaluation_without} and the latter is covered in Sec.~\ref{sec:evaluation_with}.

\begin{table}[!t]
  \centering
  \caption{ANOVA table for \nrsfm{} reconstruction error without missing data with sources  as defined in \eqref{eq:anova}. All factors are statistically significant at a 0.0005 level except $ms_{jk}$ and $mp_{jl}$.}
  \label{tab:anova_full}
  \begin{tabulary}{\linewidth}{R | L R L R L }
    \toprule
    Factor & Sum Sq.& DoF & Mean Sq. & F & p-value\\  \midrule

    $a_i$ & $3.6{\times}10^{5}$ &  15  & $2.4{\times}10^{4}$ & 204.8 & $5.5{\times}10^{-242}$ \\
    $m_j$ & $1.1{\times}10^{4}$ &  1   & $1.1{\times}10^{4}$ &  90.4 &  $3.2{\times}10^{-20}$ \\
    $s_k$ & $1.0{\times}10^{5}$      &  4  &  $2.6{\times}10^{4}$ & 219.0 & $3.6{\times}10^{-121}$ \\
    $p_l$ & $1.5{\times}10^{4}$ &  5   &  $3.0{\times}10^{3}$  &  25.6 &  $9.3{\times}10^{-24}$ \\
    $as_{ik}$ & $4.1{\times}10^{4}$ &  60  & $6.9{\times}10^{2}$ &  5.9  &  $2.9{\times}10^{-33}$ \\
    $ap_{il}$ & $4.1{\times}10^{4}$ &  75  & $5.5{\times}10^{2}$ &  4.7  &  $2.3{\times}10^{-28}$ \\
    $ms_{jk}$ & $1.3{\times}10^{3}$ &  4   & $3.2{\times}10^{2}$ &  2.7  &          0.03          \\
    $mp_{jl}$ & $1.8{\times}10^{3}$ &  5   & $3.6{\times}10^{2}$ &  3.1  &         0.0086         \\
    $sp_{kl}$ & $1.1{\times}10^{4}$ &  20  & $5.7{\times}10^{2}$ &  4.9  &  $2.3{\times}10^{-11}$ \\ \midrule
    Error & $8{\times}10^{4}$  & 689  & $1.2{\times}10^{2}$ &       &                        \\
    Total & $7{\times}10^{5}$  & 878  &                     &       &                        \\ \bottomrule
  \end{tabulary}

\end{table}

\begin{table}[!t]
  \caption{Linear term $\mu + a_i$ sorted in ascending numerical order, this is the average error for the given algorithm. Algorithms are referred to by their alias in Table~\ref{table:methods}. All numbers are given in millimeters.}
  \label{table:full_linear_algo}

 \begin{tabulary}{\linewidth}{L L L L}
 \toprule
    \textbf{MultiBody} & \textbf{KSTA} & \textbf{RIKS}\\
    29.36 & 31.94 & 32.21\\ \midrule
    \textbf{CSF2} & \textbf{MetricProj} & \textbf{CSF}\\
    32.83 & 34.09 & 41.19\\ \midrule
    \textbf{Bundle} & \textbf{PTA} & \textbf{F-Consensus}\\
    46.66 & 46.80 & 53.17\\ \midrule
    \textbf{ScalableSurface} & \textbf{CMDR} & \textbf{EM PPCA}\\
    53.88 & 53.91 & 59.21\\ \midrule
    \textbf{SoftInext} & \textbf{BALM} & \textbf{MDH}\\
    61.94 & 66.34 & 70.34\\ \midrule
    \textbf{Compressible} & \textbf{SPFM} & \textbf{Consensus}\\
    79.18 & 85.34 & 94.61\\
    \bottomrule
  \end{tabulary}

\end{table}

\subsection{Evaluation without missing data}
\label{sec:evaluation_without}
In the following, we discuss the results of the ANOVA without taking 'missing data' into account, using the model as in Eq. \eqref{eq:anova} without terms related to $d_n$:
\begin{align}
  y = \mu& + a_i + m_j + s_k + p_l + as_{ik}\label{eq:full_anova} \\
          & + ap_{il} + ms_{jk} + mp_{jl} + sp_{kl}. \nonumber
\end{align}

\begin{table}[!t]
 %\rowcolors{2}{gray!20}{white}
 \centering
 \caption{Interaction term $\mu + a_i + s_k + as_{ik}$. This is equivalent to the algorithms average error on each animatronic. Lowest error for each animatronic is marked with bold text. Algorithms are referred to by their alias in Table~\ref{table:methods}. All numbers are given in millimeters.}
 \label{tab:full_algo_ani}
 \includegraphics[width=0.49\textwidth]{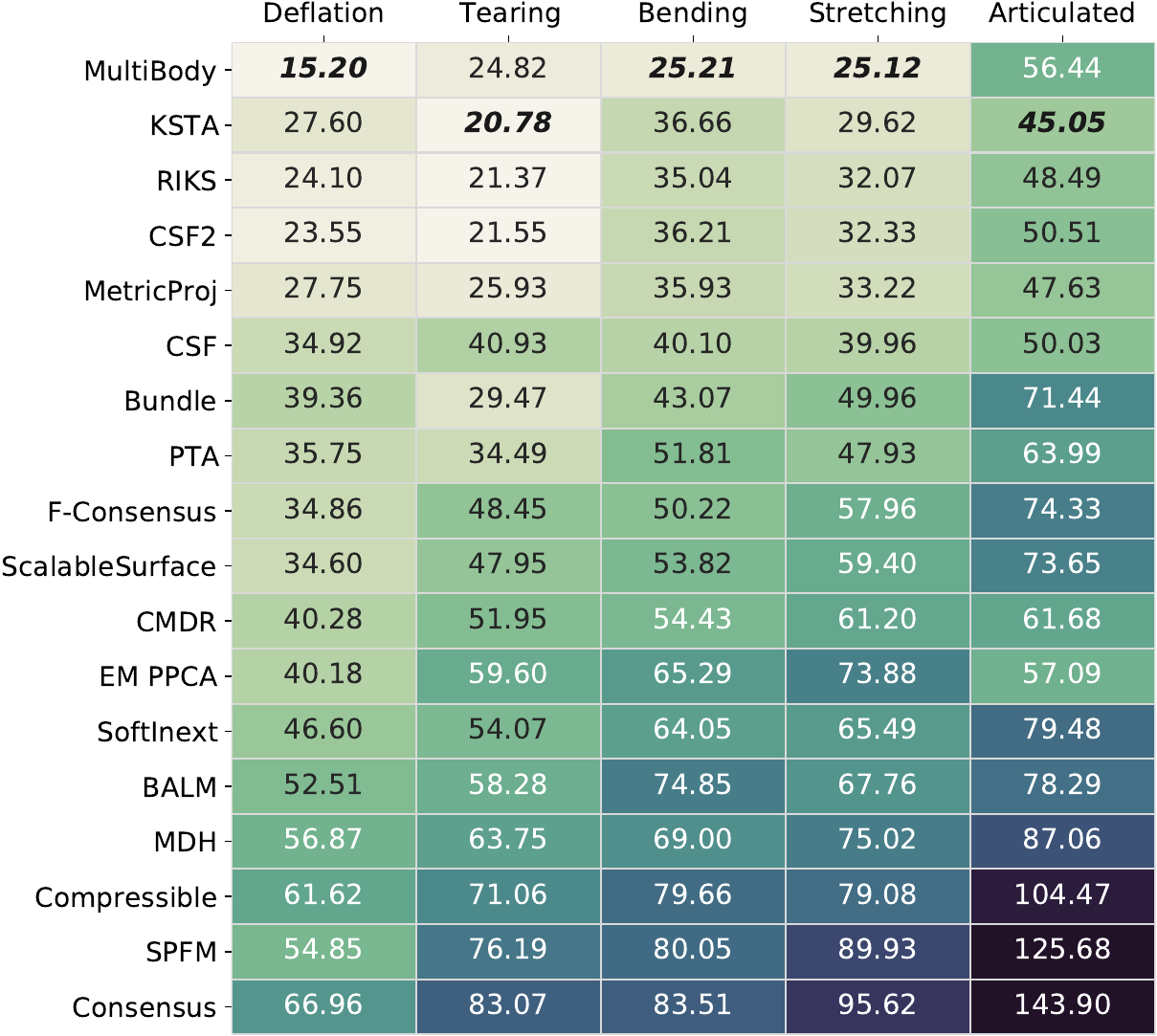}
 \centering
 \caption{Interaction term $\mu + a_i + p_l + ap_{il}$. Algorithms are referred to by their alias in Table~\ref{table:methods}. All numbers are given in millimeters.}
 \label{tab:full_algo_path}
 \includegraphics[width=0.49\textwidth]{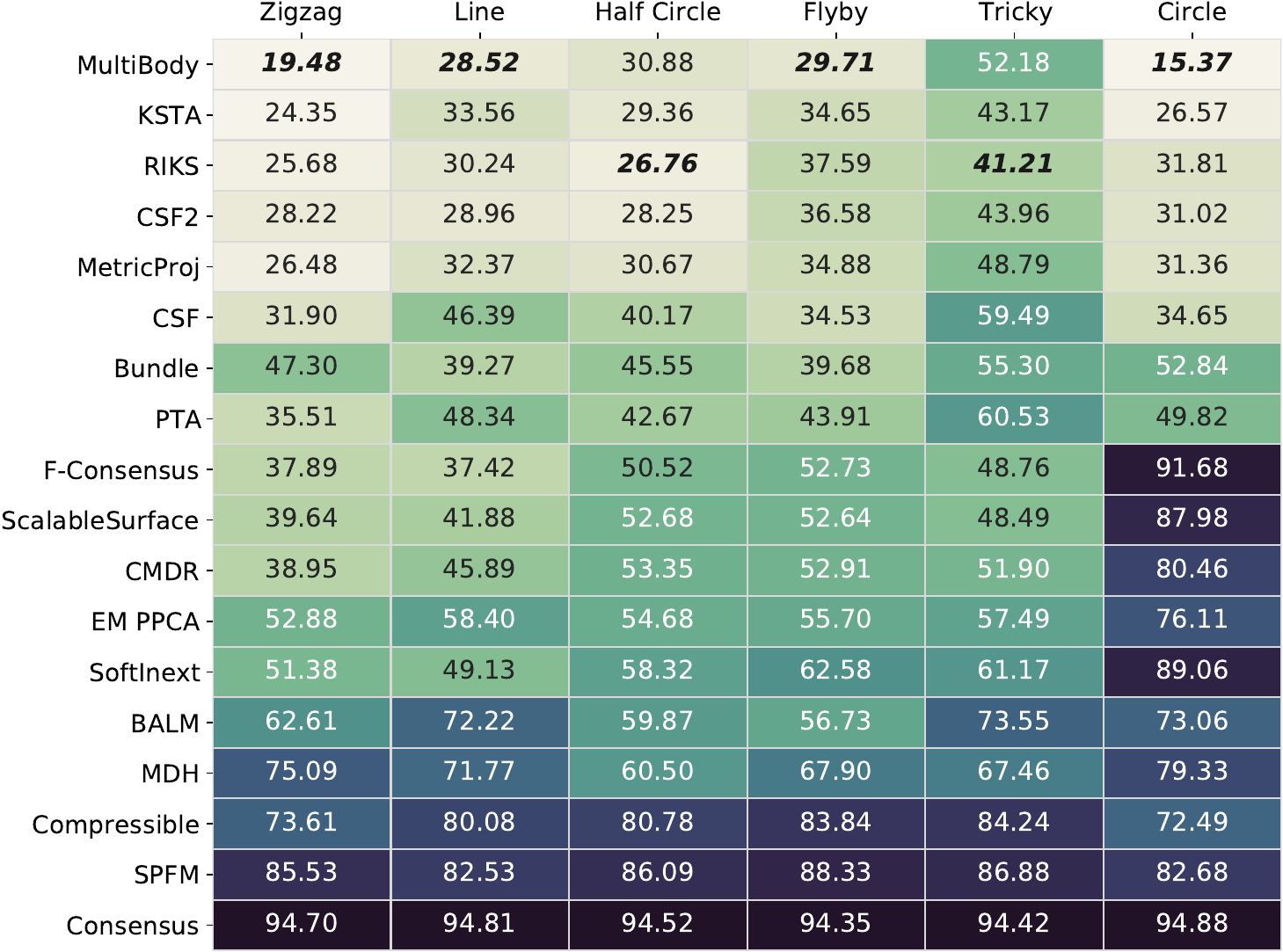}

\end{table}

The results of the ANOVA using Eq. \eqref{eq:full_anova} is summarized in Table~\ref{tab:anova_full}. All factors except $ms_{jk}$ and $mp_{jl}$ are statistically significant. As such, we can conclude that all the aforementioned factors have a significant influence on the reconstruction error. Therefore, we will explore the specifics of each factor in the following, starting with 'algorithm'.

Table~\ref{table:full_linear_algo} shows the average reconstruction error for each algorithm. The method MultiBody~\cite{MultiBody2017} has the lowest average reconstruction error over all experiments followed by  KSTA~\cite{GotardoICCV2011} and RIKS~\cite{gotardo:ECCV2012}. For more detailed insights refer to Table~\ref{tab:full_algo_ani} showing the 'algorithm' vs 'animatronic' effect on the reconstruction error. As it can be seen, MultiBody~\cite{MultiBody2017} does not have the lowest error for all animatronics, as e.g.\ KSTA~\cite{GotardoICCV2011} has a significantly lower error on the Tearing and Articulated deformations. Both of these can roughly be described as rigid bodies moving relative to each other, and it would seem KSTA~\cite{GotardoICCV2011} is the best at handling these deformations.

Methods with a physical prior, like MDH~\cite{chhatkuli2017inextensible} and SoftInext~\cite{Vicente:etal:2012} have in general lower performance, as it is evident from Tables \ref{table:methods},~\ref{tab:full_algo_ani}~and~\ref{tab:full_algo_path}. MDH~\cite{chhatkuli2017inextensible} is designed with an isometry prior, therefore one would expect it to perform well in the bending deformation. Indeed, while its interaction term $as_{ik}$ has its lowest value for the bending deformation, denoting the fitness of the chosen prior, the average reconstruction error is higher. On a more careful inspection of the reconstructed 3D sequences, it is evident that for a few frames MDH and SoftInext struggle to obtain an accurate 3D reconstruction and this affects the whole evaluation. Moreover, the 3D reconstruction shows intermittent sign flips of the 3D reconstructed shape. To this end, a stronger temporal consistency may help to reduce this negative effect and improve the method performance.

\begin{table}[!t]
  \centering
  \caption{Linear term $\mu + m_j$ sorted in ascending numerical order, this is the average error for the given camera model. All numbers are given in millimeters.}
  \label{tab:full_model}
  \begin{tabulary}{\linewidth}{C C}
  \toprule
    \textbf{Orthographic} & \textbf{Perspective}\\
    50.45 & 57.66\\ \bottomrule
  \end{tabulary}

\end{table}

\begin{table}[!t]
  \centering
    \caption{Linear term $\mu + s_k$ sorted in ascending numerical order, this is the average error for the given animatronic. All numbers are given in millimeters.}
    \label{tab:full_ani}
  \begin{tabulary}{\linewidth}{CCC}
  \toprule
    \textbf{Deflation} & \textbf{Tearing} & \textbf{Bending}\\
    39.86 & 46.32 & 54.38\\ \midrule
    \textbf{Stretching} & \textbf{Articulated}\\
    56.42 & 73.29\\ \bottomrule
  \end{tabulary}

\end{table}

\begin{table}[!t]
  \centering
  \caption{Linear term $\mu + p_l$ sorted in ascending numerical order, this is the average error for the given camera path. All numbers are given in millimeters.}
  \label{tab:full_path}

  \begin{tabulary}{\linewidth}{CCC}\toprule
    \textbf{Zigzag} & \textbf{Line} & \textbf{Half Circle}\\
    47.29 & 51.21 & 51.42\\ \midrule
    \textbf{Flyby} & \textbf{Tricky} & \textbf{Circle}\\
    53.29 & 59.94 & 61.18\\ \midrule
  \end{tabulary}

\end{table}

A similar trend can be observed in Table~\ref{tab:full_algo_path}, which shows the 'algorithm' vs 'camera path' effect on the reconstruction error. While MultiBody~\cite{MultiBody2017} has the lowest average error, it is surpassed in the Half Circle and Tricky 'camera path' by RIKS~\cite{gotardo:ECCV2012}. On the other hand, MultiBody has the lowest error under the Circle path by quite a significant margin.

From this analysis we can conclude that MultiBody performs the best on average, but is surpassed w.r.t.\ to certain camera paths and animatronic deformations by algorithms such as RIKS~\cite{gotardo:ECCV2012} and KSTA~\cite{GotardoICCV2011}. This also clearly indicates that one needs to control for both deformation type and camera motion in future \nrsfm{} comparisons, as the above conclusion could be changed by choosing the right combination of camera path and deformation. On the other hand, these findings show that \nrsfm{} performance can be optimized by choosing the right camera path (e.g. Zigzag) and the right algorithm for the deformation in question.

The camera model and its path have a significant impact on reconstruction error, a trend that can be observed from Table~\ref{tab:full_algo_path}. 

Table~\ref{tab:full_path} shows that there is a significant difference in average error w.r.t. 'camera path'. It is interesting to note, that the Circle path has one of the highest average errors, only surpassed by the Tricky camera path. The latter was specifically designed to be challenging, as such, it is surprising to find that the Circle and Tricky path's average error only differ by 3.08mm. In fact, MultiBody~\cite{MultiBody2017} seems to be the only method that benefits from the circle type of camera path, as can be seen in Table~\ref{tab:full_algo_path}. Table~\ref{tab:full_model} shows the average error of reconstructions for an orthographic and a perspective camera model. As it can be seen, there is a difference of 7.20mm, which is significant but not as large as the difference w.r.t.\ 'algorithm' (Table~\ref{table:full_linear_algo}) or 'camera path' (Table~\ref{tab:full_path}). This suggests that, while the error increases the state-of-the-art in \nrsfm{} can still operate under a perspective camera model. This is quite interesting as most \nrsfm{} approaches are not designed with a perspective camera in mind. It would seem that an orthographic or weak-perspective camera acts a reasonable approximation given the perspective distortions and the scale of the object deformation.

There is also a significant difference between the average reconstruction error of each animatronic which Table~\ref{tab:full_ani} shows. Articulated has by far the highest average reconstruction error, making it the most difficult to reconstruct for the current state-of-the-art in \nrsfm{}. Since most approaches use low-rank methods, a highly structured motion such as an Articulated is difficult to handle with a low-rank prior, especially if points are densely sampled on all joints. On the other hand, Deflation seems to be quite easy to handle for most of the state-of-the-art methods.

\begin{table}[!t]
  %\begin{table}[!t]
  \centering
    \caption{ANOVA table for \nrsfm{} reconstruction error with missing data. Factors are as defined in \eqref{eq:anova} and described at the beginning of this section. All factors are statistically significant at a 0.0005 level except $ms_{jk}$, $mp_{jl}$ and $md_{jn}$.}
    \label{tab:res_anova_miss}
  \begin{tabulary}{\linewidth}{R| L R L R L}
  \toprule
     Factor & Sum Sq.& DoF & Mean Sq. & F & p-value\\  \midrule
     $a_i$ & $1.3{\times}10^{5}$ &  8 & $1.6{\times}10^{4}$ &  90.9 & $7.7{\times}10^{-108}$ \\
     $m_j$ & $1.4{\times}10^{4}$ &  1  & $1.4{\times}10^{4}$ &  81.6 &  $1.2{\times}10^{-18}$ \\
     $s_k$ & $7.5{\times}10^{4}$ &  4  & $1.9{\times}10^{4}$ & 106.5 &  $3.8{\times}10^{-73}$ \\
     $p_l$ & $4.1{\times}10^{4}$ &  5  & $8.2{\times}10^{3}$ &  47.0 &  $8.8{\times}10^{-43}$ \\
     $d_n$ & $1.6{\times}10^{4}$ &  1  & $1.6{\times}10^{4}$ &  89.8 &  $2.7{\times}10^{-20}$ \\
     $as_{ik}$ & $1.6{\times}10^{4}$ &  32 &  $5.0{\times}10^{2}$  &  2.9  &  $3.4{\times}10^{-7}$  \\
     $ap_{il}$ & $5.6{\times}10^{4}$ &  40 & $1.4{\times}10^{3}$ &  8.0  &  $6.4{\times}10^{-37}$ \\
     $ad_{in}$ & $1.1{\times}10^{4}$ &  8  & $1.3{\times}10^{3}$ &  7.5  &  $1.1{\times}10^{-9}$  \\
     $ms_{jk}$ & $2.6{\times}10^{3}$ &  4  & $6.5{\times}10^{2}$ &  3.7  &         0.0052         \\
     $mp_{jl}$ & $2.5{\times}10^{3}$ &  5  & $5.1{\times}10^{2}$ &  2.9  &          0.013         \\
     $md_{jn}$ & $2.9{\times}10^{2}$ &  1  & $2.9{\times}10^{2}$ &  1.6  &           0.2          \\
     $sp_{kl}$ & $2.7{\times}10^{4}$ &  20 & $1.4{\times}10^{3}$ &  7.8  &  $6.7{\times}10^{-21}$ \\
     $sd_{kn}$ &$3.6{\times}10^{3}$ &  4  & $8.9{\times}10^{2}$ &  5.1  &         0.00048        \\
     $pd_{ln}$ &$8.1{\times}10^{3}$ &  5  & $1.6{\times}10^{3}$ &  9.3  &  $1.4{\times}10^{-8}$  \\ \midrule
     Error & $1.4{\times}10^{5}$ & 824 & $1.8{\times}10^{2}$ &       &                        \\
     Total & $5.7{\times}10^{5}$ & 962 &                     &       &                        \\ \bottomrule
  \end{tabulary}
%\end{table}

\end{table}

\subsection{Evaluation with Missing Data}
\label{sec:evaluation_with}
As previously mentioned, we are interested in 'missing data' and its effect on \nrsfm{}. We, thus, here use Eq. \eqref{eq:anova}, which is used to evaluate the subset of methods capable of handling missing data, as shown in Table~\ref{table:methods}. 

It should be noted that while MDH~\cite{chhatkuli2017inextensible} is nominally capable of handling missing data, it has not been included in this part of the study. The reason being that the code provided only reconstructs frames with minimum ratio of visible data, thus our error metric cannot be applied. As such, we have 9 methods in total in this category.

We treat 'missing data' as a categorical factor having two states: with or without missing data. This is because the missing percentage of our occlusion-based missing data is dependent on the 'animatronic', 'camera path' and 'camera model' factors. Additionally, there is a significant sampling bias in the occlusion-based missing data. For example, in-plane motion, like Articulated and Tearing, rarely get a missing percentage above 25\% and more volumetric motion such as Deflation rarely go below 40\% missing data. This would make it difficult to distinguish between the influence of the 'missing data' factor and the animatronic factor.

The results of the ANOVA is summarized in Table~\ref{tab:res_anova_miss} and all factors except $ms_{jk}$, $mp_{jl}$ and $md_{jn}$ are statistically significant. This means that 'missing data' has a significant influence on the reconstruction error. Table~\ref{tab:miss_algo} shows the interaction between 'algorithm' and 'missing data'. As expected, the mean error without missing data is very similar to the averages in Table~\ref{table:full_linear_algo} with KSTA~\cite{GotardoICCV2011} having the lowest expected error. However, with missing data, MetricProj~\cite{DelBue:Agapito:IJCV2011} actually has a lower average reconstruction error. This is due to its low increase in error of 5.85mm when operating under occlusion-based missing data. In comparison, KSTA~\cite{GotardoICCV2011}, CSF2~\cite{GotardoCVPR2011} and CSF~\cite{GotardoPAMI2011} are much more unstable with average increases in error of 9.65mm, 18.15mm and 13.49mm respectively. Common among the three methods is the fact that they assume a Discrete Cosine Transform (DCT) as their prior. Indeed, we see a similar increase for ScalableSurface of 16.52mm and this method also uses a DCT basis.

These results suggest that while DCT-based approaches are quite accurate without missing data, they are not very robust when operating under occlusion-based missing data. Thus, they would likely not be very robust when applied to real-world deformations, where occlusion-based missing data is unavoidable. This indicates that future research should focus on making DCT basis methods more robust or to modify the DCT model to better generalize for 'missing data'. Finally,  BALM~\cite{DelBue:etal:PAMI2012} method exhibit some peculiar behavior as its average error actually decreases by 3.33mm, contrary to expectation. A likely cause is a different computational structure of the algorithm, since the full data case uses mainly SVD for factorisation while the missing data approach has a more elaborated algorithmic approach with manifold projections and matrix entries imputation.

\begin{table}
  \centering
    \caption{Interaction between 'camera path'/'missing data'; $\mu + p_l + d_n + pd_{ln}$. Numbers are given in milimeters.}
    \label{tab:path_miss}
    \includegraphics[width=.5\columnwidth]{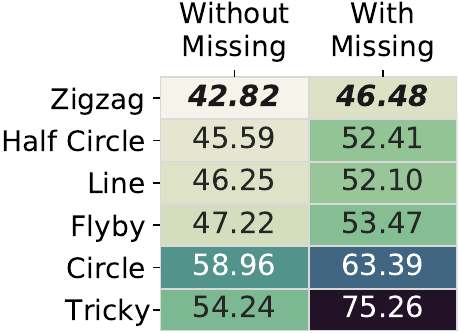}
\end{table}

\begin{table}
    \centering
    \caption{Interaction between 'animatronic'/'missing data'; $\mu + s_k + d_n + sd_{kn}$. Numbers are given in milimeters.}
    \label{tab:ani_miss}
    \includegraphics[width=.5\columnwidth]{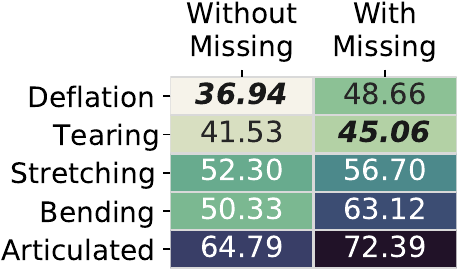}
\end{table}

\begin{table*}[!t]
  \centering
  \caption{Interaction between 'algorithm'/'missing data'; $\mu + a_i + d_n + ad_{in}$. This is the average error for each algorithm either with or without occlusion-based missing data.}
  \label{tab:miss_algo}
  \includegraphics[width=\textwidth]{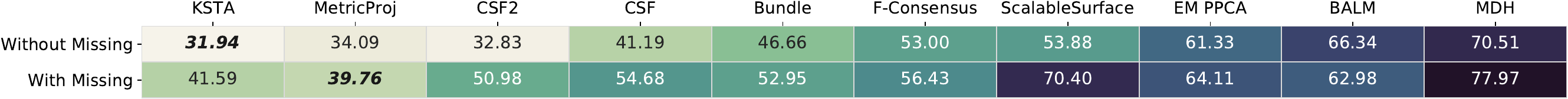}
\end{table*}

Table~\ref{tab:ani_miss} shows the average error as an interaction between 'animatronic' and 'missing data', i.e. the average reconstruction error of each animatronic with and without missing data. It is interesting to note that the in-plane deformations, i.e.\ Tearing, Stretching and Articulated, generally have a smaller increase in error with missing data compared to the more volumetric deformation, i.e.\ Deflation and Bending, compared to the error without missing data. The increase is respectively 3.96mm, 4.65mm and 8.38mm versus 12.27mm and  13.47mm. The main difference between the two groups is that the ratio of missing data is consistently low for the in-plane deformations. This would suggest that the ratio of missing data has an impact on the reconstruction error.

Table~\ref{tab:path_miss} shows the average error as interaction between 'camera path' and 'missing data'. The Tricky path has by far the highest average error. This is expected, as the small camera movement ensures that a portion of the tracked points is consistently hidden. As such, while Tricky and Circle were almost equally difficult without missing data, this is no longer the case with missing data as Circle's average error only increases by 4.9mm. Indeed, all other camera paths have approximately the same increase in error with missing data. These paths also ensure that all observed points are equally visible. What differs consistently is the spatio-temporal  distribution of missing data, which has a physical plausible structured pattern. the missing data distributions in our dataset are in contrast with previous evaluations where often missing entries were generated randomly, thus not reflecting a real 3D modelling scenario. These results also suggest that the distribution of missing data is as important as the ratio in affecting the reconstruction error. Indeed this is in line with the observations made by Paladini et al.~\cite{DelBue:Agapito:IJCV2011}.

The aforementioned observations demonstrate the importance of testing against occlusion-based missing data as it contains a spatio-temporal structure of missing data that a randomly removed subset lacks. Many \nrsfm{} methods treat missing data as a matrix fill-in problem, meaning recreating missing values from interpolation of spatio-temporally close observations. Thus, it is clear that conceptually it is much easier to interpolate random, evenly distributed missing data, compared to the spatio-temporally clustered structure of occlusion-based missing data. It is noted, that KSTA~\cite{GotardoICCV2011} and CSF~\cite{GotardoPAMI2011} were both evaluated using random subset missing data in the original works, and was found to approximately have the same performance whether from 0\% to 50\% missing data. These results are obviously quite different from the conclusion of our study and we hypothesize, that the spatio-temporal structure of our occlusion-based missing is probably the primary cause for the drop in performance of many approaches.

\section{Discussion and Conclusion} \label{sec:discussion_and_conclusion}

To summarize our findings, we would like to firstly mention that, the algorithm with the lowest error on average without missing data was found to be MultiBody~\cite{MultiBody2017}. 

There is, however, a large variation between the different algorithms performance depending on the factors chosen. As such our study does not conclude that Multibody~\cite{MultiBody2017} is definitively better than all other methods in general. As an example, for some camera paths RIKS~\cite{gotardo:ECCV2012} had lower average error than MultiBody~\cite{MultiBody2017}. Also, with missing data MetricProj~\cite{Paladini:etal:2009} has the lowest reconstruction error. Other observations include that methods with a DCT basis were found to have a great increase in error with occlusion-based missing data. In general, the  evaluated methods stay about two orders of magnitude behind the accuracy of the ground truth, showing that there is a need of improving current approaches.

Our study also shows findings that support hypotheses of where \nrsfm{} research could head in the future. Even though some of these hypotheses have been stated before in related work, the strength of our data set and evaluation is able to confirm these. Firstly, it is clear that methods using the weak perspective approximation to the perspective camera model only incur a small penalty for doing so on average. This camera model seems like a good approximation, although it should be noted, that our data set does not challenge the algorithms extremely in this regard, with only an average 1.6 fold change in the depth variations. In particular, \nrsfm{} applied in the medical domain, e.g. endoscopic imaging, may better benefit from a perspective camera model as the deforming body can be imaged at different depths while approaching with the endoscope to the regions of interest. Providing an in vivo data set for this scenario is a complex task requiring medical staff support. Some initial and promising efforts have been done for evaluating deformable registration methods \cite{modrzejewski2019vivo} that could lead to a related \nrsfm{} evaluation.  

Moreover, given continuously deforming shapes, global temporal consistency should be enforced in order to avoid frame-by-frame sign flips, reflections and other ambiguities given the stronger geometrical expressiveness of deformable models. This is truly necessary in an operative scenario where such a problem might drastically reduce the effectiveness of the \nrsfm{}  approaches.

Another main avenue of investigation was the effect of missing data. Here we found, that that this aspect has a large impact on the reconstruction error. This is somewhat at odds with previous findings, and we speculate that this has to do with our missing data having structure originating from object self occlusion, as opposed to generate missing data with random sampling. In particular, occlusion-based missing data increases the reconstruction error of all methods except BALM~\cite{DelBue:etal:PAMI2012}. Our study thus indicates this area to be a fruitful area of investigation for \nrsfm{} research.

Another observation is that the physical based methods did quite poorly compared to the methods using a statistically based deformation model. This is in a sense counter intuitive, provided that the physical models capture the deformation physics well. This, in turn, leads us to the observation that stronger efforts could be beneficial as far as better physical based deformation models. 

As stated, many of these observations, support hypothesis held in the \nrsfm{} community, and it strengths them, that we have here provided empirical support for them. On the other hand, this study also helps to validate the suitability of our compiled data set. In regard to which, it should be noted,  both deformation types and camera paths have a statistically significant impact on reconstruction error, regardless of the algorithm used. This indicated that our proposed taxonomy and the data set design has value.

All in all, we have here presented a state of the art data set for \nrsfm{} evaluation. We have applied \rev{18} different \nrsfm{} method to this data set. Methods that span the state of the art of \nrsfm{}. This evaluation validates the usability of our proposed, and publicly available data set, and gives several insights into the current state of the art of \nrsfm{}, including directions for further research.

%\begin{acknowledgements}
%\end{acknowledgements}


\bibliographystyle{spmpsci}      % mathematics and physical sciences

\bibliography{2020_ijcv_special_issue}

\end{document}